\documentclass[letterpaper]{article} 
\usepackage{aaai25}  
\usepackage{times}  
\usepackage{helvet}  
\usepackage{courier}  
\usepackage[hyphens]{url}  
\usepackage{graphicx} 
\usepackage{natbib}  
\usepackage{caption} 
\usepackage{algorithm}
\usepackage{algorithmic}

\usepackage{graphicx}
\usepackage{booktabs}

\usepackage[accsupp]{axessibility}  
\usepackage{amsmath,amsfonts}
\usepackage{array}
\usepackage[caption=false,font=normalsize,labelfont=sf,textfont=sf]{subfig}
\usepackage{textcomp}
\usepackage{url}
\usepackage{verbatim}
\usepackage{cite}
\usepackage{amssymb}
\usepackage{color}
\usepackage{listings}
\usepackage{threeparttable}
\usepackage{subfig}
\usepackage{bbding}

\usepackage{newfloat}
\usepackage{listings}
\DeclareCaptionStyle{ruled}{labelfont=normalfont,labelsep=colon,strut=off} 
\floatstyle{ruled}
\newfloat{listing}{tb}{lst}{}
\floatname{listing}{Listing}
\title{MCRL4OR: Multimodal Contrastive Representation Learning for Off-Road Environmental Perception}
\author{
    Yi Yang\textsuperscript{\rm 1},
    Zhang Zhang\textsuperscript{\rm 1},
    Liang Wang\textsuperscript{\rm 1}\\
}
\affiliations{
    \textsuperscript{\rm 1}Chinese Academy of Sciences, Institute of Automation, NLPR\\

    yangyi2021@ia.ac.cn, zzhang@nlpr.ia.ac.cn, wangliang@nlpr.ia.ac.cn
}

\usepackage{bibentry}

\begin{document}

\maketitle

\begin{figure*}[!t]
	\centering
	\includegraphics[width=\textwidth]{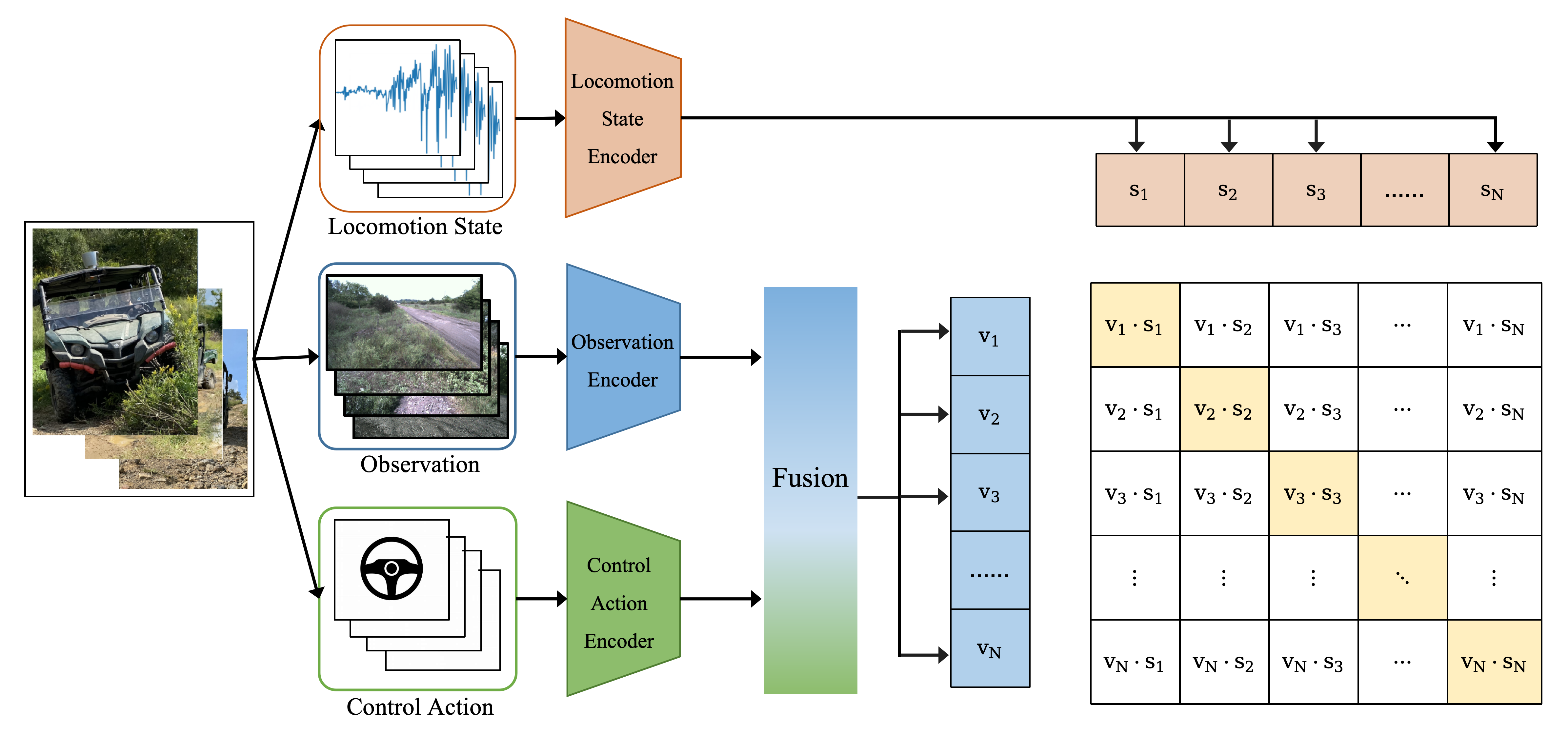}%
	\caption{Overall framework of MCRL4OR, which jointly learn a visual observation encoder, a control action encoder, and a locomotion state encoder with the goal of predicting the correct correspondence relationships in a batch of multimodal training samples.}
	\label{fig:Overall_framework}
    \vspace{-0.5cm}
\end{figure*}

\begin{abstract}
Most studies on environmental perception for autonomous vehicles (AVs) focus on urban traffic environments, where the objects/stuff to be perceived are mainly from man-made scenes and scalable datasets with dense annotations can be used to train supervised learning models. By contrast, it is hard to densely annotate a large-scale off-road driving dataset manually due to the inherently unstructured nature of off-road environments. In this paper, we propose a Multimodal Contrastive Representation Learning approach for Off-Road environmental perception, namely MCRL4OR. This approach aims to jointly learn three encoders for processing visual images, locomotion states, and control actions by aligning the locomotion states with the fused features of visual images and control actions within a contrastive learning framework. The causation behind this alignment strategy is that the inertial locomotion state is the result of taking a certain control action under the current landform/terrain condition perceived by visual sensors. In experiments, we pre-train the MCRL4OR with a large-scale off-road driving dataset and adopt the learned multimodal representations for various downstream perception tasks in off-road driving scenarios. The superior performance in downstream tasks demonstrates the advantages of the pre-trained multimodal representations. The codes can be found in \url{https://github.com/1uciusy/MCRL4OR}.
\end{abstract}

\section{Introduction}
\label{sec:intro}

In the realm of autonomous driving, it is critical to accurately perceive the surrounding environment for safe and comfortable navigation. The majority of researches focuses on urban traffic environments, where the detection and segmentation of traffic participants (e.g., pedestrians, vehicles) or traffic elements (e.g., road surface, lane lines, and traffic signs) have been widely studied. Remarkable improvements have been achieved due to the advance of deep learning technologies and the availability of large-scale and densely annotated datasets (e.g., nuScenes \cite{nuscenes}). By contrast, environmental perception in off-road driving scenarios remains a big challenge due to the unstructured and open-world nature. Accurate perception of off-road environments is crucial for autonomous vehicles (AVs) or robots to robustly and safely operate in natural (non-man-made) environments, such as \emph{harvesting in agricultural lands} and \emph{Mars exploration}. It is essential to investigate the issue of off-road environmental perception.

However, the task settings of off-road environmental perception vary with different object categories of interests and granularity of annotations. For example, Off-Road Freespace Detection (ORFD) \cite{ORFD} aims to classify the off-road visual scene into traversable areas, non-traversable areas, and unreachable areas. Nevertheless, for a harvesting robot, the above coarse-grained taxonomy is not enough to distinguish different vegetation of agricultural lands. Moreover, the definition of traversable areas is also ambiguous depending on different task requirements. Due to the huge diversity and complexity, it is challenging to collect a unified scable dataset with dense manual annotations of all-inclusive information in off-road driving scenarios. The difficulty hinders the learning of generalized representations of off-road scenes based on traditional supervised learning methods. Whereas current AVs can record the off-road driving trajectories automatically with a rich source of multimodal data (such as cameras, Lidars and inertial sensors) as well as data loggers for control actions like braking and accelerating. Thus, it is highly valuable to adopt self-supervised pre-training methods to explore the intrinsic patterns in these multimodal trajectories and learn transferable multimodal representations for various perception tasks in off-road driving scenarios.

Inspired by the success of multimodal vision-language models, we propose a multimodal contrastive representation learning approach for off-road environmental perception tasks, namely MCRL4OR, to jointly learn informative and transferable representations for visual perception, control actions, and locomotion states during off-road driving. As shown in Fig.\ref{fig:Overall_framework}, the MCRL4OR consists of three branches including a visual observation encoder, a control action encoder, and a locomotion state encoder, respectively, The objective of contrastive learning is to predict the correct correspondences between different modalities within a batch of multimodal training examples. In particular, the vision branch and control branch are early fused before the alignment with the locomotion branch. This alignment strategy indicated as $(observation+action) \leftrightarrow locomotion$ can be explained by the causal relationship that the locomotion state is the result of taking a certain control action (such as accelerating or steering) under the current landform/terrain condition (e.g., rough or smooth road surface) perceived by visual sensors. For example, when traversing an unpaved rough road with holes and cracks, the accelerating operation can lead to the ego-vehicle shaking or vibrating over the road. 


In experiments, we firstly pre-train the MCRL4OR model on the TartanDrive dataset \cite{TartanDrive}, which is the largest real-world, multimodal, off-road driving dataset. Then, a set of down-stream evaluation tasks are performed to demonstrate the advantages of the pre-trained representations, namely cross-modal retrieval, dynamics prediction, and scene segmentation. Experimental results demonstrate the effectiveness of the alignment strategy of  $(observation+action) \leftrightarrow locomotion$. Moreover, the pre-trained MCRL4OR model can consistently improve all three down-stream tasks, which validates the generalization capability of the learned multimodal representations.

The contributions can be summarized as follows. 
\begin{enumerate}
\item[(1)] We propose MCRL4OR, a multimodal contrastive representation learning approach for off-road environmental perception, which jointly learns the encoders of visual images, control actions and locomotion states with a multimodal alignment paradigm.
\item[(2)] We perform in-depth investigations into the MCRL4OR model architecture and design an optimal alignment strategy based on the intrinsic dynamics of vehicle traversal across various landform and terrain conditions.
\item[(3)] Extensive experiments are conducted to validate the advantages of MCRL4OR, including the use of the largest multimodal off-road dataset for model pre-training and evaluating the learned encoders with three down-stream tasks such as cross-modal retrieval, dynamics prediction and scene segmentation.
\end{enumerate}

\section{Related Work}
\subsection{Environmental perception for AVs}
Numerous studies have been conducted for autonomous driving in urban environments. A number of large-scale and densely annotated datasets, such as KITTI \cite{kitti}, Waymo \cite{Waymo}, Argoverse \cite{argoverse1,argoverse2}, nuScenes \cite{nuscenes} and BLVD \cite{BLVD}, have been released for the scalable training of supervised learning models. Significant advancements are achieved in various tasks such as 3D object detection \cite{bevformer}, scene segmentation \cite{bev-seg}, and occupancy prediction \cite{voxformer}, within urban traffic scenes. 

Due to the unstructured nature and wide diversity of off-road environments, current studies on off-road environmental perception are largely based on self-collected datasets for specific tasks, such as object detection or traversable region segmentation. For instance, NERC \cite{NERC} claims that the success of person detection in urban scenes cannot be directly transferred to agricultural environments and thus collects a specific dataset on person detection in agricultural scenes. Besides that, several datasets on off-road scene segmentation such as DeepScene \cite{DeepScene}, YOCR \cite{YOCR}, RUGD \cite{RUGD}, and RELLIS-3D \cite{RELLIS-3D}, have been proposed, where various modalities, from RGB images to multispectral images and 3D point clouds, have been exploited to deal with the difficulties in off-road environments. However all the above segmentation datasets are not compatible for each other due to the different annotation granularities. For example, the ORFD \cite{ORFD} dataset is proposed for traversability analysis, which segments a visual scene into three coarse-grained categories, i.e., traversable, non-traversable, and unreachable areas. While the RELLIS-3D \cite{RELLIS-3D} annotates 20 categories of objects and terrain classes, e.g., mud, grass, and rubble piles. Thus, designing a unified annotation ontology that meets the requirements of various applications in off-road scenarios is challenging.

Recently some researchers have also attempted to learn off-road perception models, guided by the decision-making and planning goals of AVs. Kahn et al. \cite{BADGR} propose a model-based reinforcement learning method to enable a wheeled robot to complete navigation tasks on off-road terrain, which leverages hours of multimodal sensor data collected through random explorations to train a perception module that understands the traversability of various types of terrain. Triest et al. \cite{TartanDrive} collect a large-scale multimodal dataset, termed TartanDrive, to learn a dynamics prediction model for future trajectory planning of an all-terrain vehicle (ATV). Their work proves that the dynamics of terrain is essential for robust and safe driving in off-road environments. 
Seo et al. \cite{Self-superviese-auxiliary-loss-in-traversability} leverage historical driving data to autonomously generate labels indicating high traversability for areas traversed by a vehicle. These self-generated labels serve as the foundation for training a neural network to identify safe-to-navigate terrains through a singular-class classification task.
Ye, et al. \cite{M2F2-ORFD-another-baseline} design a proficient multimodal learning network, named M2F2-Net, aiming to detect free space within off-road environments. Tremblay, et al. \cite{dynamics-prediction-in-unreal-engine} exploit a multimodal recurrent state-space model to integrate multi-modal learning with latent time-series prediction.

\subsection{Multimodal contrastive learning}
Contrastive learning has recently become a popular self-supervised representation learning method in computer vision (CV) community, which aims at encoding augmented versions of the same sample close to each other while pushing away those embeddings from different instances. The current dominant contrastive learning methods include SimCLR \cite{SimCLR, SimCLRv2} and MoCo \cite{MoCo, MoCov2}. 

Contrastive learning can be naturally extended for multimodal representation learning, where each unimodal component is considered as one augmented version of a multimodal sample. For example, CLIP \cite{CLIP} has been proposed for scalable vision-language pre-training, which jointly learns an image encoder and a text encoder to predict the correct alignment relationships in a batch of image and text training pairs. The similar idea has been adopted to other modalities such as video \cite{videoCLIP}, audio \cite{audioCLIP}, 3D point cloud \cite{PointCLIP}, and inertial measurement unit (IMU) \cite{IMU2CLIP}. In this work, we also adopt a CLIP-like multimodal contrastive alignment paradigm to learn the representations of visual images, control actions, and locomotion states for off-road environmental perception. Meanwhile, the contrastive learning has become a popular way for multi-modal learning in autonomous systems. For example, Ma, et al. \cite{COMPASS} propose a multimodal pre-training approach that needs to optimize three contrastive losses based on spatial connections, temporal connections, and spatiotemporal connections between spatial and temporal modalities. Zhang, et al. \cite{youtube-contrastive} present an action-conditioned contrastive learning method to jointly learn visual features and policy features. However, it fails to consider the interactions between the visual observations, actions, and locomotion states, which are crucial in off-road driving scenarios. 

\begin{figure*}[!t]
	\centering
	\includegraphics[width=\textwidth]{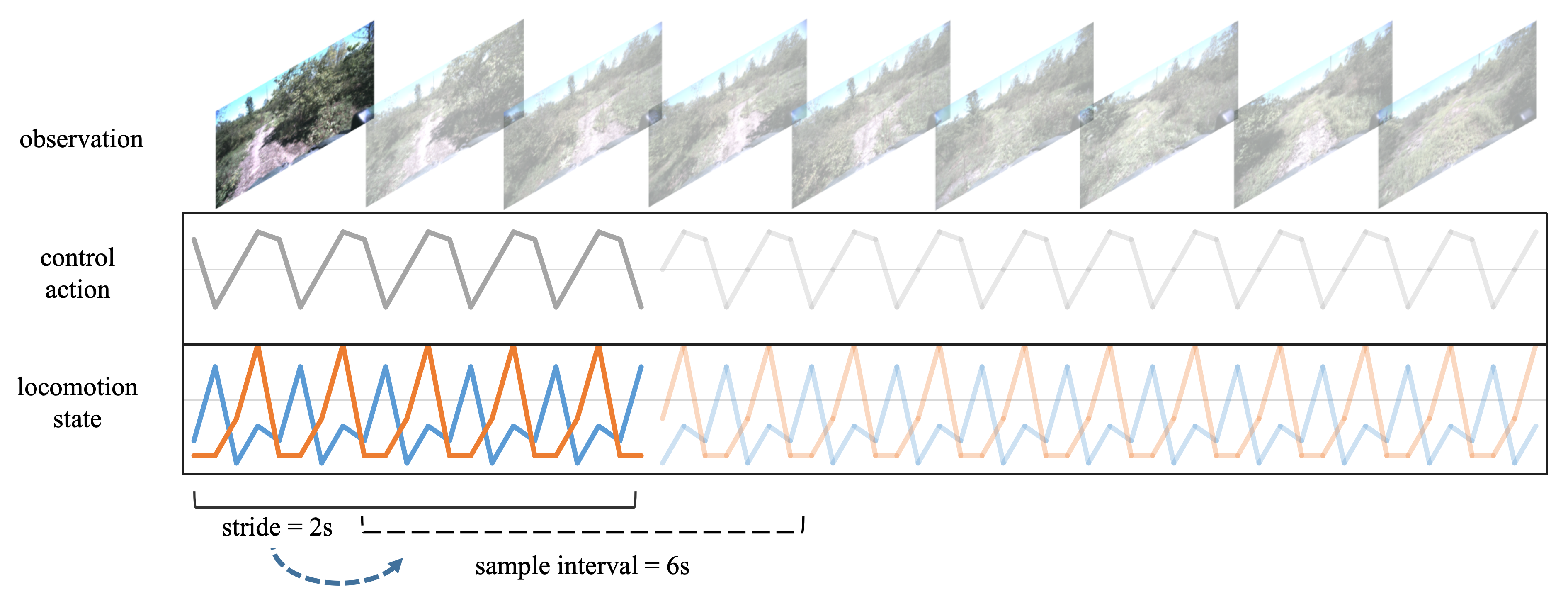}
	\caption{Construction diagram of triplet samples, where a sample consists of a frame of visual image and its following 6s (240 frames) time series of locomotion states and control actions. Every adjacent samples have a 2 second stride.}
	\label{fig:data_preparation}
    \vspace{-0.5cm}
\end{figure*}

\section{Methods}
\subsection{Multimodal Data Categories}
We assume an off-road AV (ego-agent) has been mounted with multiple perception sensors (e.g., camera and IMU) to perceive its surroundings and data loggers to record control actions (throttle and steering). Then the multimodal trajectories gathered by these sensors and data loggers are categorized into three groups: \emph{observations}, \emph{locomotion states}, and \emph{control actions}.

\begin{itemize}
    \item\textbf{Observations} denoted as $O$ are the visual/depth data captured by exteroceptive sensors, e.g., cameras or lidars, which describe the exteroceptive situation of the ego-agent in the off-road environment. In our work, observations consist of visual images captured by a stereo camera recording terrain information ahead of the ego agent.
    
    \item\textbf{Locomotion states} denoted as $S$ are the proprioceptive states of the ego-agent, including various locomotion measurements such as wheeled odometers, acceleration, angular velocity, and force transducer. The locomotion states reflect the ego-agent's perception of traversing different terrains in the off-road environment. In this work, each locomotion state contains 1) ego-pose, which is the position vector $p = (x, y, z)$ and quaternion orientation $q = (q^x, q^y, q^z, q^w)$; 2) ego-motion, which includes the inertial data (angular velocity and linear acceleration), shock travel and wheel revolutions per minute (RPM). The channel width of a locomotion state is 27. Detailed information about each channel can be found in the Supplementary Material.
  
    \item\textbf{Control actions} denoted as $C$ are the driver's actions to control vehicles moving in off-road environments, which are represented as a two-dimensional vector $(\mu^t,\mu^s)$ corresponding to throttle and steering, respectively. We adhere to the definitions in TartanDrive \cite{TartanDrive}, where the throttle action $\mu^t$ ranges from 0 to 1, with 1 corresponding to full throttle. The steering action takes a value between -1 and 1, with -1 representing a hard left turn and 1 representing a hard right turn.
\end{itemize}

By the above definitions, the ego-agent's perception state in the current off-road environment can be represented by a pair of exteroceptive state ($O$) and proprioceptive state ($S$). 

\subsection{Data Preparation}
Before introducing the details of MCRL4OR, the procedure for data preparation is outlined here. This process synchronizes all sensors to a unified frequency, and then splits the trajectories of $O$, $S$, $C$ to produce a set of triplet samples for training the MCRL4OR.

\begin{itemize}
    \item\textbf{Frequency synchronization.}
    Various sensors often work at different frequencies, for instance, in the TartanDrive \cite{TartanDrive}, the IMU operates at 200Hz (400Hz in their released dataset), while the wheel RPM sensor reads at 50Hz, and 10Hz for the stereo camera. As previously mentioned, the locomotion state consists of multiple high-frequency sensor data stream such as IMU and wheel RPM readings. Working with raw data would entail a significant amount of time spent on aligning the timestamps from different sensors for the purpose of loading training samples. To mitigate this issue, we perform an explicit synchronization process, where all locomotion sensors and the control actions are upsampled to a common frequency of 400Hz for a subtle temporal alignment. Then, an average downsampling operation is performed with a unified frequency of 40Hz to reduce the data redundancies and noises. Finally, the frequency of the stereo camera observations is preserved at 10Hz.
    \item\textbf{Construction of triplet samples.}  
    We need to select an appropriate granularity to split the continuous trajectories into a scalable and diverse set of triplet samples for training the MCRL4OR model. Unlike urban traffic scenes, vehicles in off-road environments often move at relative slower speeds. To ensure that the vehicle can drive through the off-road area observed by the camera, we sample the trajectories of $O$, $S$, $C$ with a temporal window of 6 seconds to construct a triple sample $<o_i, s_i, c_i>$, where $i$ is the sample index. For each sample, $o_i$ is chosen as the first image frame in the sampled temporal window from trajectory $O$. $s_i$ corresponds to all 240 frames in the 6 seconds from trajectory $S$ with a channel width of 27, i.e., a $240\times 27$ tensor. $c_i$ also contains 240 frames with 2 channels, i.e., a $240\times 2$ tensor. Finally, we sample the triplet samples every 2 seconds, ensuring that two adjacent samples have no more than 4 seconds overlapping period. Fig. \ref{fig:data_preparation} illustrates the process of constructing triplet samples.
\end{itemize}
In this work, we adopt the largest multimodal off-road driving dataset, i.e., TartanDrive \cite{TartanDrive}, to pre-train the MCRL4OR. After data preparation, a total of 61470 triplet samples are collected from 630 trajectories in the TartanDrive.

\subsection{MCRL4OR Framework}
As presented in Fig. \ref{fig:Overall_framework}, the proposed MCRL4OR framework is inspired by the success of CLIP framework \cite{CLIP} for vision-language pre-training. Here, for a mini-batch of $n$ training samples $<o_i, s_i, c_i>_{i=1}^{n}$, the MCRL4OR adopts three encoders to extract the features of each input triplet, i.e., $v^o_i=f^o(o_i)$, $v^s_i=f^s(s_i)$, $v^c_i=f^c(c_i)$. Specifically, the observation encoder is based on the Swin-T \cite{Swin}, where the classification head is replaced by a 2-layer multilayer perceptron (MLP). As for the locomotion state encoder $f^s(*)$ and the control action encoder $f^c(*)$, the sequence encoder proposed in IMU2CLIP \cite{IMU2CLIP} is adopted, which is a stacked 1D-CNN with GroupNorms \cite{GroupNorm}.

Before the multimodal alignment between the observations and the locomotion states, the features of observation $v^o_i$ and control action $v^c_i$ are fused by a 2-layer MLP-based fusion module to acquire a joint feature $v^m_i =mlp(v^o_i, v^c_i)$. The reason behind this early fusion strategy is as follows. \emph{The locomotion state is the proprioceptive sense of the ego-agent when traversing across the current terrain (depicted by the visual observation) with a certain control action (e.g., accelerating or steering)}. The alignment strategy denoted as $(observation+action) \leftrightarrow locomotion$ can be explained as the learning of visual affordance \cite{gibson79} in off-road environments, i.e., the ego-agent learns to predict the possible proprioceptive sense based on the given control action and the observation of the current terrain.

Then, for each pair of features $v^s_i$ and $v^m_j$, the similarity is calculated using cosine similarity.
\begin{equation}
sim(v^s_i,v^m_j) = \frac{v^s_i}{\lVert v^s_i\rVert_2}\cdot \frac{v^m_j}{\lVert v^m_j\rVert_2},
\label{eq:1}
\end{equation}
where $i,j \in \{1...n\}$.

Finally, a $n \times n$ similarity matrix is obtained to calculate the contrastive alignment loss for each mini-batch of training samples.

\subsection{Contrastive Loss}
Similar to the loss function in the CLIP \cite{CLIP}, the goal is to contrast the paired features $v^s_i$ and $v^m_i$ from the same source with that from different sources in the current mini-batch. For a locomotion state feature $v^s_i$, the training loss function can be summarized as:
\begin{equation}
    l_{(v^s_i,*)}=-\frac{\exp(sim(v^s_i,v^m_i)/\tau)}{\sum_{j=1}^n \exp(sim(v^s_i,v^m_j)/\tau)}
\end{equation}
where $\tau$ is a learnable temperature parameter.

Since the alignment relationship is symmetric, there exists a symmetric loss function for a given joint feature $v^m_j$:

\begin{equation}
l_{(*,v^m_i)}=-\frac{\exp(sim(v^s_i,v^m_i)/\tau)}{\sum_{j=1}^n \exp(sim(v^s_j,v^m_i)/\tau)}
\end{equation}

The overall loss function is defined as follows.
\begin{equation}
L=\sum_{i=1}^n\left(l_{(v^s_i,*)}+l_{(*,v^m_i)}\right)
\end{equation}

\section{Datasets and Evaluation Tasks}

\subsection{Datasets}

Two large-scale off-road datasets, i.e., TartanDrive \cite{TartanDrive} and ORFD \cite{ORFD}, are adopted to validate the effectiveness of the proposed MCRL4OR.

TartanDrive dataset \cite{TartanDrive} is built for the task of off-road dynamics prediction, which collects 630 trajectories with roughly 200,000 driving interactions on a modified Yamaha Viking ATV with 7 unique sensing modalities in a variety of terrain including tall grass, rocks, and mud. This is the current largest real-world multimodal off-road driving dataset, both in terms of the number of interactions and sensing modalities. Following the training/test dataset splits in \cite{TartanDrive}, we use 46101 triplets sampled from the trajectories in the training set to pre-train the MCRL4OR model. Then, two evaluation tasks, i.e., cross-modal retrieval and off-road dynamics prediction, are performed on the test set. 

\begin{figure}
	\centering
	\includegraphics[width=\linewidth]{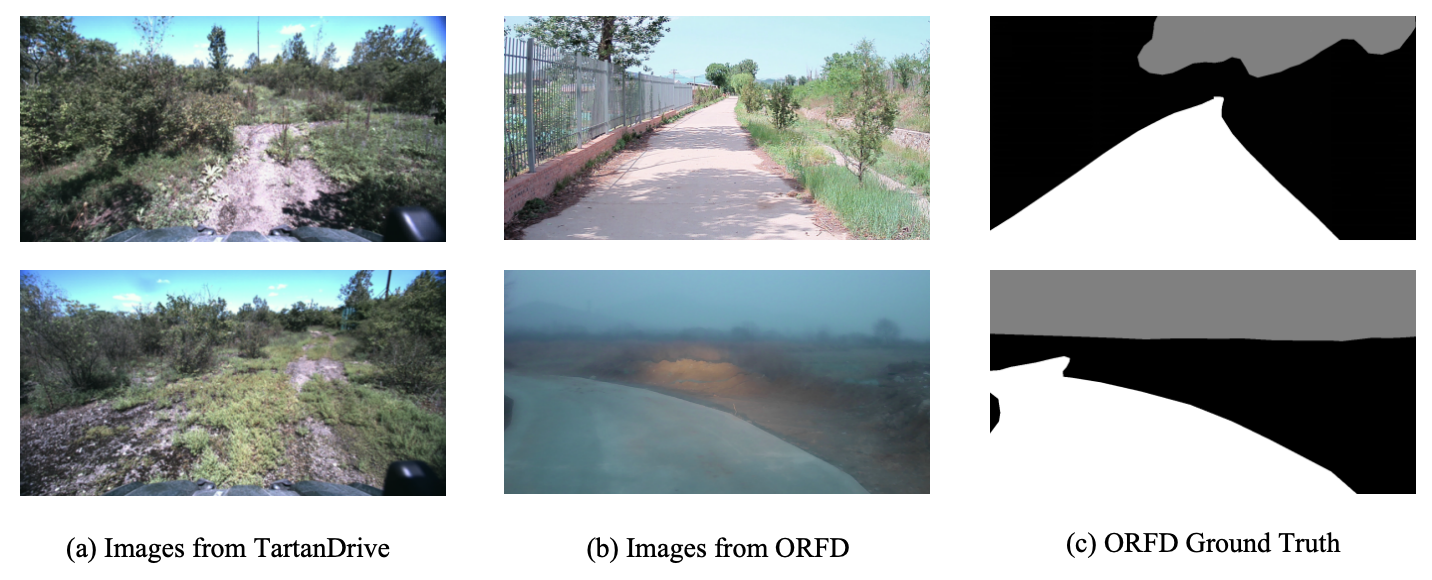}%
	\caption{Some off-road image samples in our work. (a) the images from TartanDrive \cite{TartanDrive}. (b) the images from ORFD \cite{ORFD}. (c) the annotations of semantic segmentation in ORFD, white area for traversable, black area for non-traversable and gray for non-reachable.}
	\label{fig:ORFD_samples}
    \vspace{-0.5cm}
\end{figure}

ORFD dataset \cite{ORFD} is proposed for traversability analysis in off-road environments. It collects 30 sequences covering a distance of about 3km in off-road environments with various scenarios and weather conditions. A total of 12198 Lidar point cloud and RGB images are annotated at pixel level with three classes, i.e., traversable, non-traversable, and unreachable area. In this work, we utilize the ORFD dataset to showcase the generalization capabilities of the pre-trained MCRL4OR model in the task of off-road semantic segmentation. Some image samples and annotations are shown in Fig. \ref{fig:ORFD_samples}.

\begin{figure*}
	\centering
    \includegraphics[width=1\textwidth]{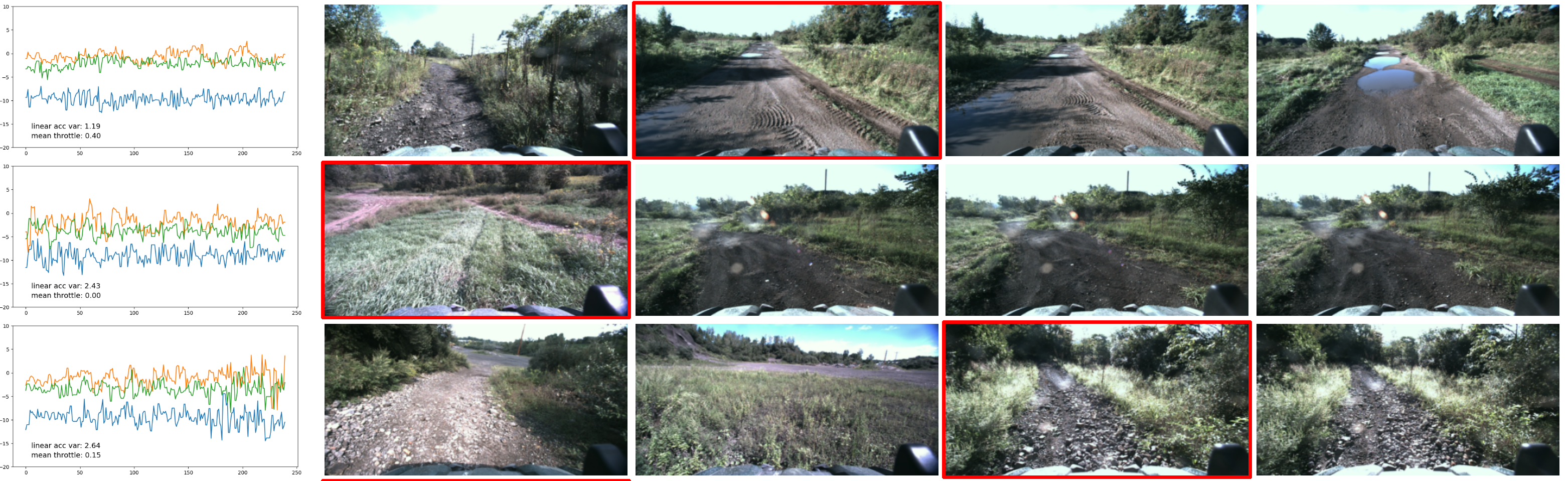}
	\caption{The left column is linear accelerations along three axes in locomotion state. The blue, orange and green lines are corresponded to linear acceleration along z, y and x axes. Right are top 4 similar observations retrieved. The paired positive image are bounded by a red frame.}
	\label{fig:retrieve}
    \vspace{-0.5cm}
\end{figure*}

\subsection{Evaluation Tasks}
As mentioned above, three tasks are adopted to evaluate the advantages of the pre-trained MCRL4OR.

\textbf{Task 1: Cross-modal retrieval} is adopted to evaluate the effects of multimodal alignments of the pre-trained MCRL4OR, which has been commonly used in the studies of multimodal self-supervised learning \cite{IMU2CLIP,CLIP,videoCLIP}. Here, we conduct the evaluation task on the test set of TartanDrive. For each triplet sample, bi-directional retrievals i.e., $locomotion \rightarrow (observation+action)$ and $(observation+action) \rightarrow locomotion$ are performed. The evaluation metric is the ranked accuracy (rank-n Acc.), which measures the percentage of targets retrieved within the top-k ranked results over all queries. Notably, for each query, there is only one target from the same triplet sample.

In this task, two baselines are tested for comparison. The first is a random retrieval, which serves as a lower-bound performance. The second one is to use the simple alignment strategy $locomotion \rightarrow observation$ for pre-training MCRL4OR, which ignores the intervention of control actions. In this case, the input of control action input is padded with zeros to train the MCRL4OR without the input of control actions. This baseline is adopted to investigate alternative architecture for the MCRL4OR model.
    
\textbf{Task 2: Off-road dynamics prediction} is to predict the vehicle's trajectory and orientation given the planned control actions and history trajectories. TartanDrive \cite{TartanDrive} proves that in unstructured environments, the dynamics of the terrain is a significant factor in affecting the ego-vehicle's trajectory following driving actions. We adhere to the same dynamics prediction setup as in TartanDrive and construct two dynamics prediction models using two backbones, i.e., Gate Recurrent Unit (GRU) \cite{GRU} which has been used in \cite{TartanDrive} and Informer \cite{Informer} which is a more advanced transformer based predictor. For both backbones, the pre-trained locomotion states encoder is fixed to extract locomotion features. The evaluation metric is rooted mean square error (RMSE) of the predicted locomotion state (x-y-z coordinates and quaternion).


In comparison, two baselines in \cite{TartanDrive} are adopted. The first is the kinematic bicycle model (KBM), which is a baseline of mathematical deterministic iterative methods. The other is the original GRU based model in \cite{TartanDrive}.


\textbf{Task 3: Off-road semantic segmentation} aims to segment the traversable areas in off-road environments, which is used to demonstrate the generalization capability of the pre-trained MCRL4OR model in the ORFD dataset \cite{ORFD}. ORFD treats segmentation as a pixel-level classification task, using accuracy (acc), precision, recall, f1 and mIoU to evaluate  model's performance. Concretely, we replace the image encoder of the OFF-Net \cite{ORFD}, with the observation encoder of the pre-trained MCRL4OR model. Then, we follow the train/validation/test set split in the ORFD to fine-tune the OFF-Net. Since the observation encoder in the MCRL4OR model, i.e., Swin-T, is more advanced than that in OFF-Net, we conduct an additional MCRL4OR pre-training with the image encoder in OFF-Net for a fair comparison. Due to the space limitations, the details can be found in the Supplementary Materials.

Besides the OFF-Net, the performance of other baselines, e.g., FuseNet \cite{fusenet} and SNE-RoadSeg \cite{sne_roadseg}, are also reported.

\section{Experiments}
In this section, we will introduce the implementation details in experiments, and the results of various evaluation tasks and the ablation studies.

\subsection{Implementation Details}
As for pre-training MCRL4OR, Adam \cite{Adam} optimizer with recommended hyper-parameters is used. The total training epochs are 20. The learning rate is linearly increased from $10^{-4}$ to $10^{-3}$ in the first 10 epochs, then kept unchanged for the last 10 epochs. The batch size is set to 64 samples per GPU. The output feature dimensions for all encoders are 128. Following the studies \cite{SimCLR, SimCLRv2, MoCo, MoCov2}, a two-layer MLP is set to project encoder features to a common alignment feature space. Additionally, a two-layer MLP is used for fusing observation features and control action features before the alignment with locomotion state features. To enforce the model to focus on the ground, we crop the images to use only the lower half as input. The MCRL4OR is pre-trained on a server (Ubuntu 22.04) with 6 GPUs (Nvidia Titan X, 12GB Memory).

For the task of \textbf{cross-modal retrieval}, we report the rank-1, rank-10, and rank-50 accuracies. A retrieval is deemed successful if the positive paired samples lies in the top-k ranking results.

For the task of \textbf{off-road dynamics prediction}, Adam \cite{Adam} optimizer is used for training the dynamic prediction model. The batch size in fine-tunning is set to 64. Instance normalization \cite{instance_norm} is used to normalize the input of locomotion states. To address the non-stationary nature of the locomotion state, we predict the trajectory's differentiation between two adjacent frames instead of absolute trajectory. During evaluation, we accumulate the differentiations to obtain the predicted positions. The pre-trained locomotion state encoder is frozen and extracts historical locomotion state features as the initial hidden state of GRU, or as the input tokens of the encoder of Informer. The train/test split keeps the same as the setting in \cite{TartanDrive}.

For the task of \textbf{off-road semantic segmentation}, the input image size is set to 640 × 352 and the batch size is set to 12. All other settings are kept the same settings as the ORFD \cite{ORFD}. The performance of the last model after 30 training epochs is reported.

\subsection{Results}



\subsubsection{Cross-Modal Retrieval} 

\begin{table}[!t]
    \setlength{\abovecaptionskip}{0cm}
    \setlength{\belowcaptionskip}{0cm}
    \setlength{\tabcolsep}{0.2cm}
	\caption{Retrieval rank-n accuracy}
	\center
    \begin{threeparttable}
        \begin{tabular}{cccccccc}
    	\bottomrule
    	Task & method & rank-1& rank-10& rank-50\\    
        \hline
        o\& a $\rightarrow$ s & rand. & 0.02 \% & 0.12 \% & 0.53 \% \\
        o\& a $\rightarrow$ s & mask a & 0.09 \% & 1.25 \% & 5.31 \% \\
        o\& a $\rightarrow$ s & ours & 5.59 \% & 22.88 \% & 41.73 \% \\
        \hline
        s $\rightarrow$ o\& a & rand. & 0.02 \% & 0.12 \% & 0.53 \% \\
        s $\rightarrow$ o\& a & ours & 5.50 \% & 22.86 \% & 43.43 \% \\
    	\bottomrule
    	\end{tabular}
     	\begin{tablenotes}[para]
    	\footnotesize 
        \item{'rand' : random guessing method.}
        
        \item{'mask a' : action is masked in both pre-training and retrieval phases, e.g., the simple strategy of $observation \leftrightarrow locomotion$ }
        
        \end{tablenotes}
    \end{threeparttable}
	\label{table:state_retrieve_observation}
    \vspace{-0.2cm}
\end{table}


The rank-n accuracy of cross-model retrieval are reported in Tab. \ref{table:state_retrieve_observation}. Compared to the proposed alignment strategy (\emph{ours}), the simple alignment strategy (\emph{mask a}) has a drastic drop in retrieval performance on the test set, e.g., 41.73\% vs. 5.31\% in the rank-50 group. This exhibits a severe over-fitting of representations learned by the simple strategy, although the observation and locomotion state can be well-aligned in training phase.

Besides the quantitative results, we visualize two representative locomotion-based queries with different acceleration distributions and the corresponding top 4 retrieved images. As shown in Fig.\ref{fig:retrieve}, the locomotion in the first row presents a lower variance of 3-axes accelerations. The retrieved images that the vehicle is driving through a puddle of water are reasonable, as the wet soil is soft and the vehicle often moves slowly in such cases. As for the queried locomotion with higher variance in the second row, the retrieved images also show a rough uneven road surface covered with obstacles like small rocks. The visualization results demonstrate that the MCLR4OR model can learn generalized representations of observations and locomotion states reflecting the dynamics of various terrains.

\subsubsection{Off-Road Dynamics Prediction}
The results of dynamic prediction are shown in Table \ref{table: overall dynamics prediction}, where the RMSE denotes the errors of the predicted position and orientation. The results of baselines including GRU and KBM are reported from \cite{TartanDrive}. As shown in the results, the RMSE performance of GRU-based prediction model can be improved by a large margin from 16.7\% to 5.9\%, when equipped with locomotion state features extracted by the pre-trained MCRL4OR. Compared to GRU, Informer as a more advanced transformer-based prediction model can further gain a performance improvement from 5.9\% to 3.5\%. Moreover, with the pre-trained locomotion state encoder, we only need 240 epochs to fine-tune the prediction models. In contrast, the original GRU-based model from \cite{TartanDrive} requires 5000 epochs for training from scratch.     

\begin{table}[!t]
    \setlength{\abovecaptionskip}{0cm}
    \setlength{\belowcaptionskip}{0cm}
    \setlength{\tabcolsep}{0.2cm}
	\caption{Overall results of dynamics prediction}
	\center
	\begin{threeparttable}
	\begin{tabular}{c|c|c|c}
	\bottomrule
	Method & Backbone & epochs & RMSE \\
	\hline
	KBM & Mathematical model & - & 1.1638\\
	TartanDrive & GRU  & 5000 & 0.1674 \\
	\hline
	Ours & GRU & 240 & 0.0593\\
	Ours & Informer & 240 & 0.0355\\
	\bottomrule
	\end{tabular}
	\end{threeparttable}
	\label{table: overall dynamics prediction}
    \vspace{-0.2cm}
\end{table}

\subsubsection{Off-Road Semantic Segmentation}
The evaluation results of semantic segmentation on the ORFD dataset \cite{ORFD} are presented in Table \ref{Segmentation_overall}. The results of other baselines are reported according to ORFD \cite{ORFD} and M2F2\cite{M2F2-ORFD-another-baseline}. As shown in the results, using only the vision modality, the observation encoder pre-trained by MCRL4OR achieves a 6.3\% (88.8\% -> 95.1\%) increase in Acc., compared to the OFF-Net baseline \cite{ORFD}. Furthermore, with multimodal input, the Acc. can be enhanced by an additional 1\% (95.8\%->96.8\%) performance gain. Although the model M2F2 \cite{M2F2-ORFD-another-baseline} achieves a better performance due to the special design for fusing point clouds information, after we pretrain the encoder of M2F2 with our method, the Acc, F1 and mIoU are all slightly better than naive M2F2.

\begin{table}[!t]
    \setlength{\belowcaptionskip}{0cm}
    \setlength{\abovecaptionskip}{0cm}
    \setlength{\tabcolsep}{0.2cm}
	\caption{Overall results of traversability semantic segmentation}
	\center
	\begin{threeparttable}
		\begin{tabular}{c|c|c|c|c|c}
			\bottomrule
			Method & RGB & 3D Pts & Acc & F1 & mIoU\\
			\hline
			OFF-Net & \Checkmark &  & 88.8\% &  80.6\% & 67.5\%\\
			Ours & \Checkmark &  & 95.1\% &  90.8\% & 83.1\%\\
            \hline
			FuseNet & \Checkmark & \Checkmark & 87.4\% & 79.5\% & 66.0\%\\
			SNE-roadseg & \Checkmark & \Checkmark & 93.8\% & 89.6\% & 81.2\%\\
			OFF-Net  & \Checkmark & \Checkmark & 94.5\% & 90.3\% & 82.3\%\\
			OFF-Net* & \Checkmark & \Checkmark & 95.8\% & 92.4\% & 85.8\%\\
			OFF-Net+Ours* & \Checkmark & \Checkmark & 96.8\% & 94.2\% & 89.0\%\\
            M2F2*  & \Checkmark & \Checkmark & 98.1\% & 96.4\% & 93.1\%\\
            M2F2+Ours*  & \Checkmark & \Checkmark & \textbf{98.3}\% & \textbf{96.7}\% & \textbf{93.6}\%\\
            \bottomrule
			\end{tabular}
	\begin{tablenotes}
		\footnotesize
		\item {'*' means the result is from our experiments, otherwise, reported by reference paper.}
	\end{tablenotes}				
	\end{threeparttable}
	\label{Segmentation_overall}
\end{table}

Given our method is pretrained on the lower half of images, mostly grounds, hindering the model from understanding the sky, the major 'unreachable' class in segmentation. Furthermore, we crop the images and only segment lower half. These experiments can be found in supplementay materials.

\section{Conclusion}
In this paper, we propose MCRL4OR, a general-purpose pre-training approach to learning multimodal representations for supporting a variety of off-road perception tasks. The proposed alignment strategy aims to learn the intrinsic relationships between observations, locomotion states and control actions. Three downstream evaluation tasks have been performed comprehensively to validate the advantages of the pre-trained representations learned by the MCRL4OR. Experimental results show that the MCRL4OR can capture the dynamics of various terrains in off-road environments and improve the perception capability over different tasks.

\bibliography{aaai25}

\section{Reproducibility Checklist}
This paper:

\begin{itemize}
\item Includes a conceptual outline and/or pseudocode description of AI methods introduced (yes)
\item Clearly delineates statements that are opinions, hypothesis, and speculation from objective facts and results (yes)
\item Provides well marked pedagogical references for less-familiare readers to gain background necessary to replicate the paper (yes)
\end{itemize}

Does this paper make theoretical contributions? (no)

Does this paper rely on one or more datasets? (yes)

If yes, please complete the list below.

\begin{itemize}
    \item A motivation is given for why the experiments are conducted on the selected datasets (yes)
    \item All novel datasets introduced in this paper are included in a data appendix. (NA)
    \item All novel datasets introduced in this paper will be made publicly available upon publication of the paper with a license that allows free usage for research purposes. (NA)
    \item All datasets drawn from the existing literature (potentially including authors’ own previously published work) are accompanied by appropriate citations. (yes)
    \item All datasets drawn from the existing literature (potentially including authors’ own previously published work) are publicly available. (yes)
    \item All datasets that are not publicly available are described in detail, with explanation why publicly available alternatives are not scientifically satisficing. (NA)
\end{itemize}

Does this paper include computational experiments? (yes)

If yes, please complete the list below.

\begin{itemize}
    \item Any code required for pre-processing data is included in the appendix. (yes).
    \item All source code required for conducting and analyzing the experiments is included in a code appendix. (yes)
    \item All source code required for conducting and analyzing the experiments will be made publicly available upon publication of the paper with a license that allows free usage for research purposes. (yes)
    \item All source code implementing new methods have comments detailing the implementation, with references to the paper where each step comes from (NA)
    \item If an algorithm depends on randomness, then the method used for setting seeds is described in a way sufficient to allow replication of results. (NA)
    \item This paper specifies the computing infrastructure used for running experiments (hardware and software), including GPU/CPU models; amount of memory; operating system; names and versions of relevant software libraries and frameworks. (yes)
    \item This paper formally describes evaluation metrics used and explains the motivation for choosing these metrics. (no)
    \item This paper states the number of algorithm runs used to compute each reported result. (yes)
    \item Analysis of experiments goes beyond single-dimensional summaries of performance (e.g., average; median) to include measures of variation, confidence, or other distributional information. (no)
    \item The significance of any improvement or decrease in performance is judged using appropriate statistical tests (e.g., Wilcoxon signed-rank). (no)
    \item This paper lists all final (hyper-)parameters used for each model/algorithm in the paper’s experiments. (yes)
    \item This paper states the number and range of values tried per (hyper-) parameter during development of the paper, along with the criterion used for selecting the final parameter setting. (yes/partial/no/NA)
\end{itemize}

\clearpage
\begin{center}
    \LARGE\bfseries Appendices
\end{center}
\vspace{2ex}

\section{Detailed information of TartanDrive data}

Following the data description in TartanDrive's github repository, \url{https://github.com/castacks/tartan_drive}, and the paper of TartanDrive\cite{TartanDrive}, the data registered in robot operating system (ROS) bag can be categorized into:

\begin{itemize}
    \item \textbf{Action}: robot action, 2-dimensional vectors corresponding to the throttle and steering positions;
    \item \textbf{State}: robot pose, 7-dimensional vectors corresponding to x-y-z cartesian coordinate and quaternion;
    \item \textbf{RGB Image}: images of terrain ahead of the car;
    \item \textbf{RGB Map}: images of terrain ahead of the car generated from BEV view;
    \item \textbf{Height Map}: depth information of RGB Map;
    \item \textbf{IMU}: 6-dimensional vectors including angular velocity around 3 axis and linear acceleration in 3 axis;
    \item \textbf{Shock Pos}: a 4 dimensional vector indicating the vibration/shock information of four wheels;
    \item \textbf{Wheel RPM}: a 4 dimensional vector indicating the revolutions per minute of four wheels.;
    \item \textbf{Pedals}: intervention data of brake pedal, including a brake pedal position, and a boolean intervention signal that indicated when the brakes exceeded a threshold.
\end{itemize} 

Detailed ROS bag data description table is shown in Table.\ref{table:tartan_data_description}.

\section{Experiments Illustration}

Given the diverse focus of off-road autonomous driving and the novel modality we use, vision \& IMU sensors. It's not easy to directly conduct our experiments based on other's work. So we mainly develop based on the code of CLIP\cite{CLIP}, then replace the encoders we need into the framework of CLIP.

We will clarify the settings of experiments in this section by the encoders we use and the aim of our experiments.

There are four pretraining experiments for our various downstream task and ablations. The expriments are shown as in Tab.\ref{table:exp_demonstrate}. All the pretraining experiments share the same hyper-parameters and converge during training.

\begin{table}[]
    \caption{All data registered in TartanDrive's ROS bag.}
    \setlength{\tabcolsep}{0.3cm}
    \center
    \begin{tabular}{llll}
	\bottomrule
        Modality& Dimension& Frequency\\
        \hline
        State& 7& 50Hz\\
        Action& 2& 100Hz\\
        RGB Image& 1024 x 512& 20Hz\\
        RGB Map& 501 x 501& 20Hz\\
        Heightmap& 501 x 501& 20Hz\\
        IMU& 6& 400Hz*\\
        Shock Pos& 4& 50Hz\\
        Wheel RPM& 4& 50Hz\\
        Pedals& 2& 50Hz\\    	
        \bottomrule
    \end{tabular}
    \begin{tablenotes}
		\footnotesize
			\item * Though TartanDrive describes the IMU frequency is 200Hz, the data in the released github repository is recorded at the frequency of 400Hz.
	\end{tablenotes}
    \label{table:tartan_data_description}
\end{table}

\begin{table*}[]
    \caption{vision, state encoders and the experiments where they are used}
    \setlength{\tabcolsep}{0.3cm}
    \center
    \begin{tabular}{llll}
	\bottomrule
         vision encoder & state encoder & experiments\\
        \hline
        Swin-T backbone & custom conv encoder & Cross-modal retrieval, Dynamics prediction, RGB ablation of seg. \\
        Swin-T backbone & GRU & GRU for ablation of dynamics prediction in 3rd row of Tab.2 in paper \\
        OFF-Net backbone & custom conv encoder & OFF-Net encoder for segmentation in 7th row of Tab.3 in paper\\
        M2F2 backbone & custom conv encoder & M2F2 encoder for segmentation in 9th row of Tab.3 in paper \\
    \bottomrule
    \end{tabular}
    \label{table:exp_demonstrate}
\end{table*}

\section{Cross-Modal Retrieval}

\begin{figure*}
	\centering
    \includegraphics[width=\textwidth]{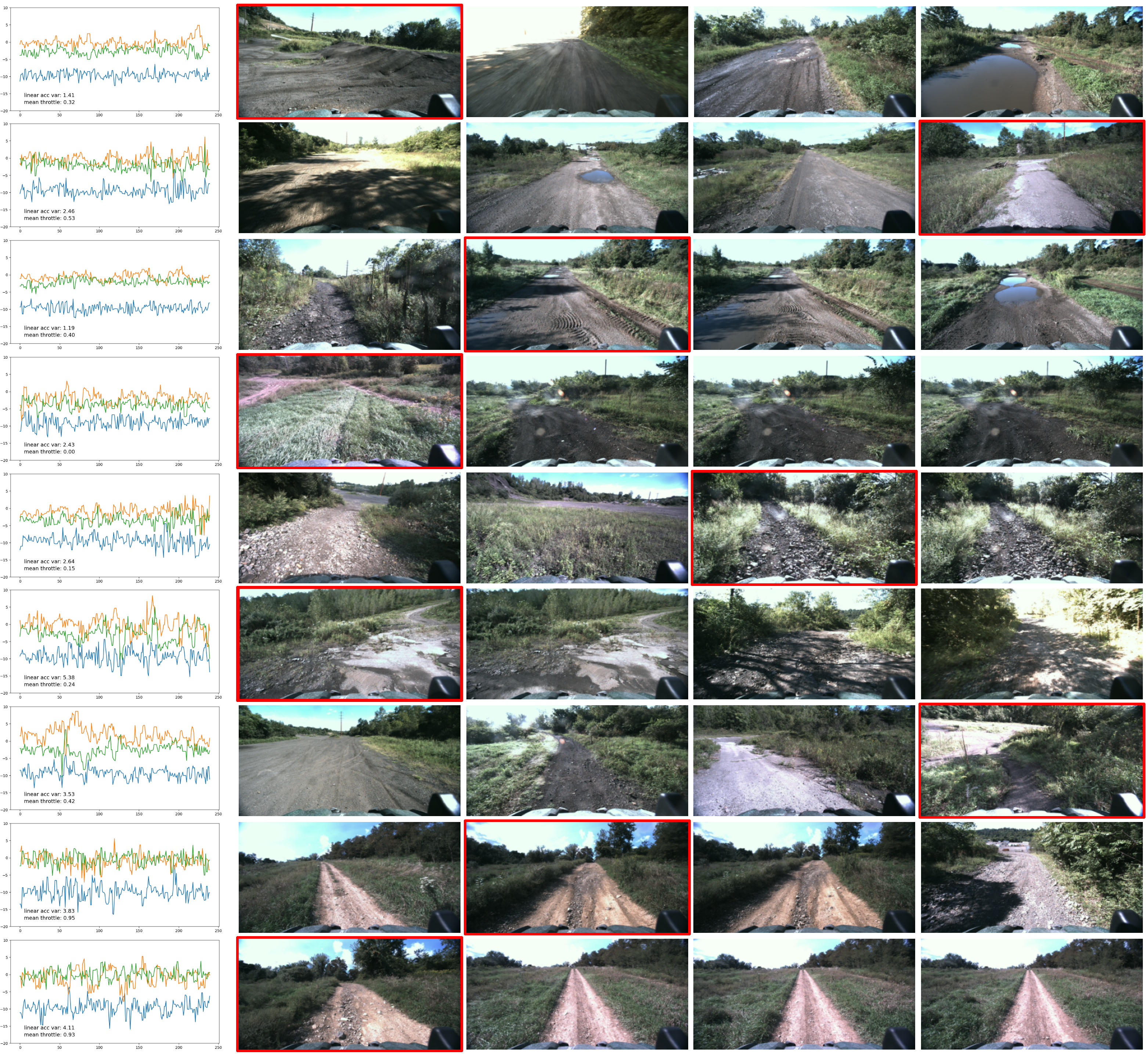}
     \caption{Left is linear acceleration along three axes in locomotion state. The blue, orange and green lines are corresponded to linear acceleration along z, y and x axes, the mean throttle is recorded along with the linear acceleration. Right are top 4 similar observations retrieved. The paired positive image are bounded by a red frame.}
	\label{fig:supp_retrieve}
\end{figure*}

More cross-modal retrieval cases are show in Fig.\ref{fig:supp_retrieve}. In general, the variance of linear acceleration is correlated to the roughness of terrain, yet the velocity and throttle also affect the driving state.



\begin{figure}
    \centering
    \includegraphics[width=\linewidth]{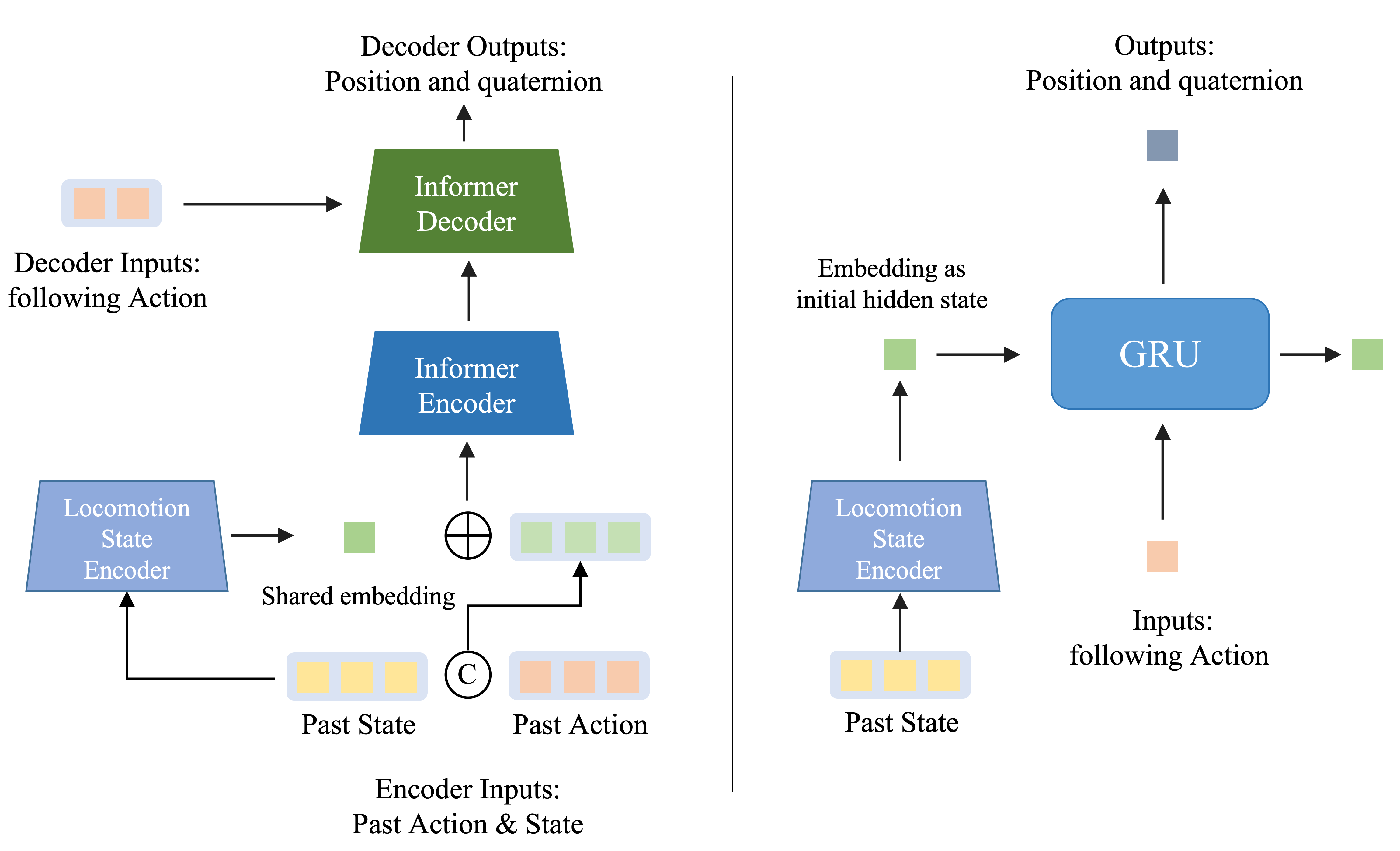}
    \caption{Diagram of locomotion state encoder in dynamics prediction. The left is how state embedding is how state embedding is used in Informer, a shared embedding added to every input token. The right is how state embedding is used in GRU, as the GRU's initial hidden state.}
    \label{fig:dynamics_pred}
\end{figure}

\section{Dynamics Prediction}

The way we use locomotion state encoder in dynamics prediction is illustrated in Fig.\ref{fig:dynamics_pred}.

GRU is an auto-regressive RNN-based model, also the baseline model in TartanDrive. It takes the action and the historical location motion state as the inputs and outputs the future position and orientation results. The locomotion state encoder in the pre-trained MCRL4OR model is adopted to generate the initial hidden state embedding of GRU from the past locomotion state.

Informer is a transformer-based encoder-decoder architecture. It also takes the past state and action as the encoder's input, and outputs all positions and orientations at once. The locomotion state encoder in the pre-trained MCRL4OR model is adopted to transfer the past locomotion state into a state embedding. Then the embedding is further added to the tokens generated from both the past state and control action.

That only state encoder is adopted in terms of two considerations: (1) only one encoder out of three encoders in MCRL4OR is utilized to avoid coupling effects; (2) Compared to the control actions and the observations of visual image, the locomotion state has more informative content for the task of dynamic prediction.

In the dynamics prediction, the pre-trained encoder is used to encode the historical $m$ seconds' states and actions. And the prediction model predicts the following $n$ seconds' position and quaternion results take the following $n$ seconds' action and output of the encoder as input and . In the TartanDrive, $m$ and $n$ are set to be 2 and 2. For simplicity of the experiment, we firstly set $m$ to be 4 and $n$ to be 2, due to the triplet samples having 6 seconds time intervals. Further, to make the experiment convincing, we set our setting to be the same as TartanDrive. The locomotion state encoder and action encoder firstly pad input to 6 seconds by replicating, thus they can process varying input length of less than 6 seconds. So the settings in the evaluation task are agnostic to the pre-training settings. After previous adjustments, the RMSE of the informer-based model increases from 0.0355 to 0.0394. A larger historical context does contribute to experiment results, yet our method still outperforms the baseline.

\section{Semantic Segmentation}

\begin{table*}[!t]
    \setlength{\abovecaptionskip}{0cm}
    \setlength{\belowcaptionskip}{0cm}
    \setlength{\tabcolsep}{0.1cm}
	\caption{Ablation on models with only RGB as input}
	\center
	\begin{threeparttable}
		\begin{tabular}{c|c|c|c|c|c|c|c}
			\bottomrule
			Encoder & ImageNet & MCRL4OR  & Acc & Precision & Recll & F1 & mIoU\\
			\hline
			OFF-Net & - & - & 88.8\% & 75.5\% & 86.4\% & 80.6\% & 67.5\% \\
			OFF-Net & - & \checkmark & 94.1\% & 95.8\% & 81.4\% & 88.0\% & 78.6\% \\
            \hline
			Swin-T \cite{Swin} & - & - & 88.0\% & 84.4\% & 67.7\% & 75.1\% & 60.2\% \\
            Swin-T & - & \checkmark & 87.2\% & 72.4\% & 84.3\% & 77.9\% & 63.8\% \\
			Swin-T & \checkmark & - & 94.4\% & 86.1\% & 94.4\% & 90.1\% & 81.9\% \\
			Swin-T & \checkmark & \checkmark & 95.1\% & 92.1\% & 89.4\%  & 90.8\% & 83.1\%\\
		\bottomrule
		\end{tabular}
	\end{threeparttable}
	\label{table:segmentation ablation rgb only}
\end{table*}

Because Swin-T \cite{Swin} is adopted as the observation encoder in MCLR4OR pre-training, which is larger and more powerful backbone compared to the visual encoder in OFF-net, we further conduct ablations with only the vision modality. Concretely, we replace the Swin-T in MCRL4OR pre-training with the visual encoder of OFF-Net, and then fine-tune the pre-trained visual encoder with ORFD dataset. As presented in the second row of Table \ref{table:segmentation ablation rgb only}, the performance is also increased significantly from 88.8\% to 94.2\% in Acc. due to the MCRL4OR pre-training. This phenomenon clearly validates the benefits of MCRL4OR pre-training for the downstream segmentation task.

Furthermore, we investigate the effects of different initialization methods for the Swin-T encoder, i.e., initialization with ImageNet-based pre-training or random initialization. The results are presented in the Swin-T group of Tab.\ref{table:segmentation ablation rgb only}. We can find that the MCRL4OR pre-training on TartanDrive consistently improves the Swin-T for both initialization methods, except for a slight drop in the Acc. when using random initialization. Meanwhile, the sole MCLR4OR pre-training on TartanDrive with randomly initialized Swin-T does not outperform the OFF-Net baseline. This indicates that the pre-training on larger-scale datasets such as ImageNet is essential when employing the advanced transformer-based models (e.g., Swin-T) with a large parameter capacity. 

\begin{figure}
    \centering
    \includegraphics[width=\linewidth]{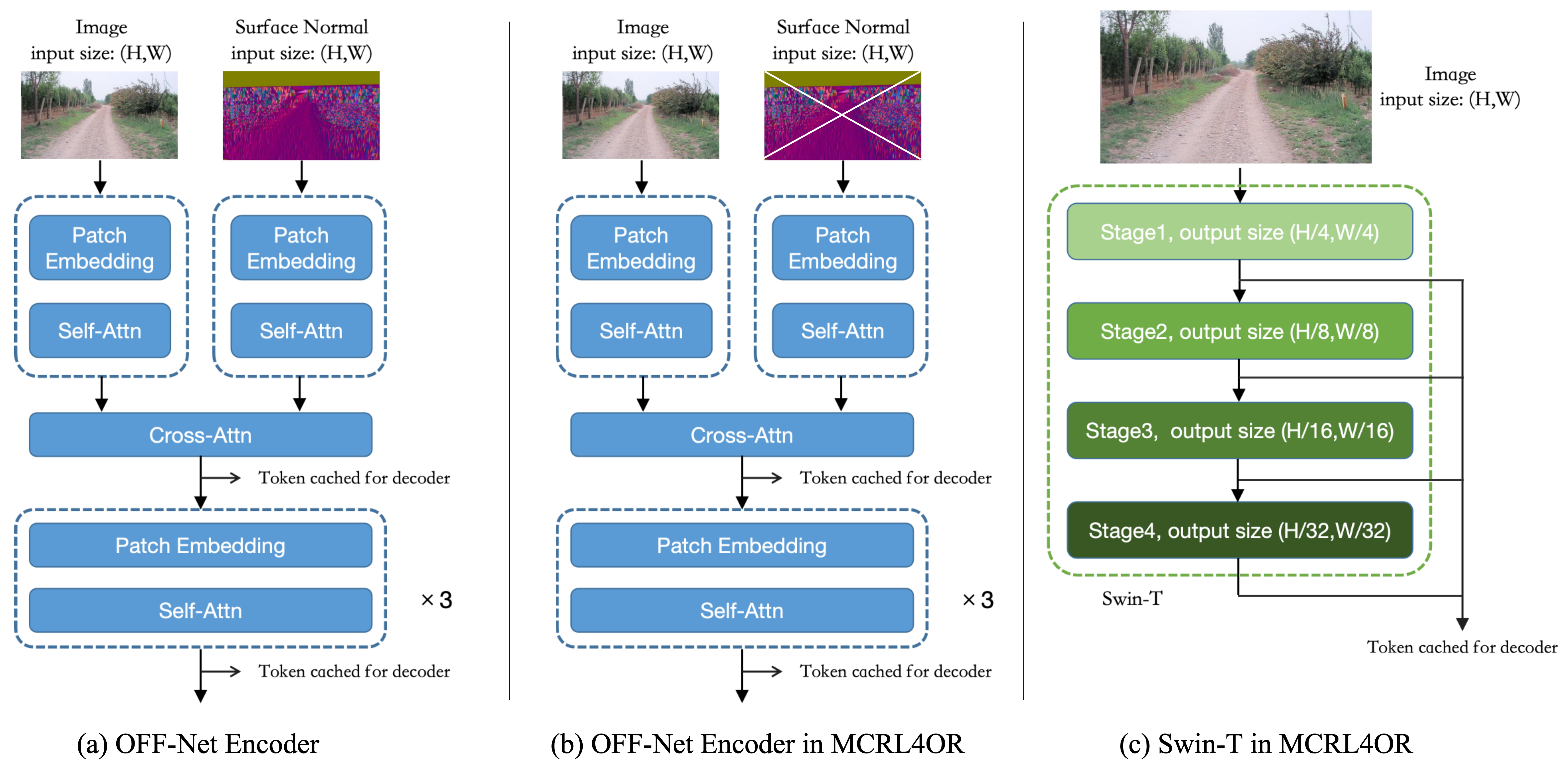}
    \caption{(a) is a simplified diagram of original OFF-Net Encoder. (b) is how OFF-Net encoder is used in MCRL4OR with the surface normal input masked. (c) is the Swin-T encoder in MCRL4OR. In MCRL4OR, only the last stage's output is used for the alignment.}
    \label{fig:OFF-Net}
\end{figure}

The OFF-Net from ORFD\cite{ORFD} is illustrated in Fig.\ref{fig:OFF-Net}. The input image and surface normal are firstly separately tokenized and associated with the multi-head self-attention (MHSA)\cite{sup_attention}, then the features from two modalities are fused by multi-head cross attention before sent to the decoder consisting of a series of N transformer layer.

\begin{figure}
    \centering
    \includegraphics[width=\linewidth]{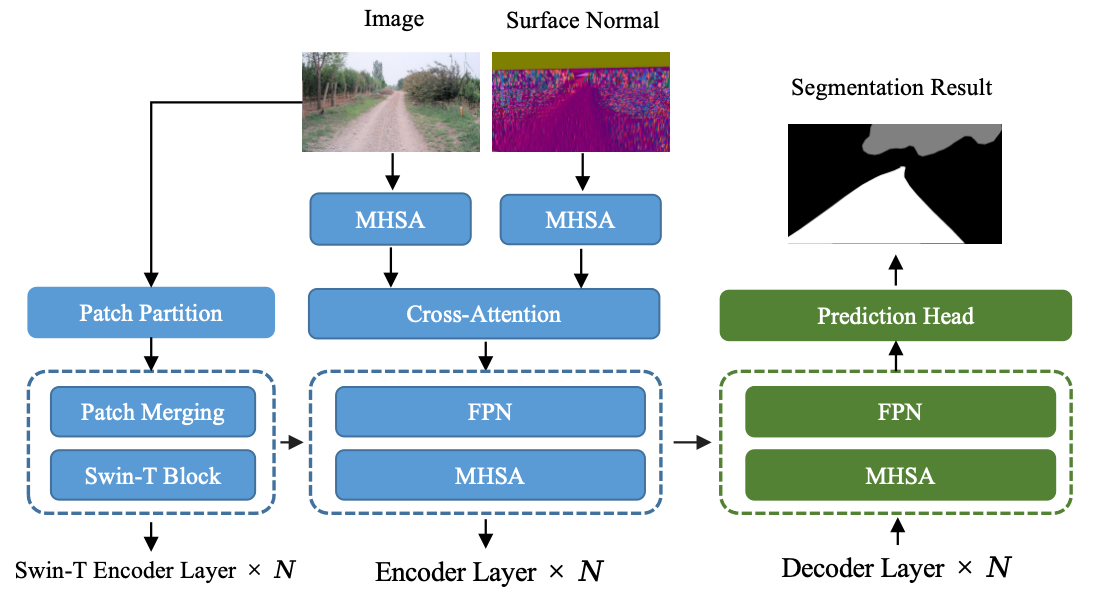}
    \caption{Diagram of segmentation with only camera as input.}
    \label{fig:OFF-Netv2}
\end{figure}

For the ablation studies that only take images as input, we use the encoders (b) and (c) in Fig.\ref{fig:OFF-Net} to replace the original encoder of OFF-Net. The former brings no extra parameters, while the latter introduces an extra Swin-T encoder. The number of parameters after the replacement is 27.24M and 54.76 respectively.

For the ablations that take both images and surface normals as input, the way we use the observation encoder is shown in Fig.\ref{fig:OFF-Netv2}. The Swin-T is utilized as a parallel feature extractor. The image feature obtained by the Swin-T encoder layer serves as the query of the decoder layer of OFF-Net.

We also compare a single-modal pre-training method i.e., SimCLR\cite{SimCLR}. An visual encoder and a locomotion state encoder with random initialization are trained by SimCLR with the Tartan's training set separately, where the encoders' architectures and hyper-parameters are identical to the MCRL4OR. In dynamics prediction, the RMSE of Informer with the pre-trained locomotion state encoder is 0.0393, inferior to 0.0355 of ours. And the segmentation results shown in Tab.\ref{tab:singl_modal_vs_multi_modal} also validate the superiority of the proposed MCRL4OR.

\begin{table}[h]
\centering
\caption{single-modal vs MCRL4OR pretrain}
\label{tab:singl_modal_vs_multi_modal}
\begin{tabular}{ccccccc}
\hline
Pretrain & Acc. & Pre. & Rec. & F1 & IoU \\ \hline
single-modal & 86.6 & 77.7 & 70.2 & 73.8 & 58.4 \\ 
MCRL4OR & 87.2 & 72.4 & 84.3 & 77.9 & 63.8  \\ \hline 
\end{tabular}
\end{table}

\section{Limitation and Discussion}

It is crucial to recognize the limitations of the current study and suggest promising avenues for future research. 

Firstly, the scalability of the TartanDrive \cite{TartanDrive} used for MCRL4OR pre-training are still limited, which might hinder the generalization capability of the learned representations. Thus, it is necessary to collect larger-scale and more diverse off-road driving datasets for more convincing model comparison.  

Secondly, the diversity of evaluation tasks remains limited. Performance saturation is observed in the tasks of dynamic prediction and semantic segmentation. It would be beneficial to delve into the capabilities of the MCRL4OR model for off-road navigation tasks, where MCRL4OR could function as a world model for a reinforcement learning (RL) agent. The agent could utilize the model to image the future locomotion states based on different actions taken when navigating off-road environments.

Finally, high-fidelity driving simulators, e.g., Carla\cite{CARLA}, have been widely used for developing autonomous driving agents within a closed-loop training and test environment. Thus, researchers would profit from an off-road driving simulator that incorporates high-fidelity visual sensing and a physics engine to model dynamics when AVs traversing diverse terrain conditions.

\end{document}